\documentclass{article}
\usepackage{spconf,amsmath,graphicx}
\usepackage{subcaption}
\usepackage{amssymb}
\usepackage{amsfonts}
\usepackage{mathtools}
\usepackage{multirow}
\usepackage{bm}
\usepackage{nicefrac}
\usepackage{tabularx}
\usepackage{booktabs}
\usepackage{pifont}
\usepackage{amsthm}
\usepackage{extarrows}
\usepackage{hyperref}
\usepackage{balance}

\newtheorem{prop}{Proposition}
\usepackage{array}

\makeatletter
\def\hlinewd#1{%
  \noalign{\ifnum0=`}\fi\hrule \@height #1 \futurelet
   \reserved@a\@xhline}
\makeatother

\title{Two-step Sound Source Separation: Training on Learned Latent Targets}
\name{Efthymios Tzinis${^\natural}$ \quad Shrikant Venkataramani${^\natural}$ \quad Zhepei Wang${^\natural}$ \quad Cem Subakan${^\flat}$ \quad Paris Smaragdis$^{\natural,\sharp}$\thanks{Supported by NSF grant \#1453104}}
\address{$^\natural$ University of Illinois at Urbana-Champaign \\
$^\flat$ Mila--Quebec Artificial Intelligence Institute\\
$^\sharp$ Adobe Research\\ }
\begin{document}
\ninept
\maketitle
\begin{abstract}
In this paper, we propose a two-step training procedure for source separation via a deep neural network. In the first step we learn a transform (and it's inverse) to a latent space where masking-based separation performance using oracles is optimal. For the second step, we train a separation module that operates on the previously learned space. In order to do so, we also make use of a scale-invariant signal to distortion ratio (SI-SDR) loss function that works in the latent space, and we prove that it lower-bounds the SI-SDR in the time domain. We run various sound separation experiments that show how this approach can obtain better performance as compared to systems that learn the transform and the separation module jointly. The proposed methodology is general enough to be applicable to a large class of neural network end-to-end separation systems. %Single-channel sound source separation has greatly benefited from modern deep learning approaches that utilize a mask-based architecture which is trained end-to-end using a time-domain loss. In essence, the mixture signal is passed through an encoder in order to get a latent representation and then the separation module estimates some masks that correspond to each sound source. The masked latent representations are translated to time-domain signals through a linear decoder. In this work, we propose a general two-step training framework of all these mask based architectures. First we train the encoder and decoder parts individually in order to find the latent representation of the mixture and the clean sources which serve as targets. In the second step we train the separation module to reconstruct these latent targets by forcing them to maximize scale-invariant signal to distortion ratio (SI-SDR) on the latent domain. We prove that this measure lower bounds the time-domain SI-SDR separation performance by  that lower bounds the  which consists of individually parts  that instead of using short-time Fourier transform (STFT)
%Recent works propose to utilize a mask 
%Still praying to Paris.
\end{abstract}
\section{Introduction}
\label{sec:intro}
Single-channel audio source separation is a fundamental problem in audio analysis, where one extracts the individual sources that constitute a mixture signal \cite{belouchrani1998blindsourceseparationTFrepresentations}. Popular algorithms for source separation include independent component analysis \cite{choi2005blindICA}, non-negative matrix factorization \cite{le2015deepNMF} and more recently supervised \cite{huang2014deep, hershey2016deepclustering, UNETS, luo2019convTasNet, wang2019deepPhaseReconSpeakerSep} and unsupervised \cite{tzinis2019unsupervised, seetharaman2019bootstrapping, drude2019unsupervised} deep learning approaches. In many of the recent approaches, separation is performed by applying a mask on a latent representation, which is often a Fourier-based or a learned domain. Specifically, a separation module produces an estimated masked latent representation for the input sources and a decoder translates them back to the time domain.

Many approaches have used the short-time Fourier transform (STFT) as an encoder to obtain this latent representation, and conversely the inverse STFT (iSTFT) as a decoder. Using this representation, separation networks have been trained using a loss defined over various targets, such as: raw magnitude spectrogram representations \cite{huang2014deep}, ideal STFT masks \cite{huang2015jointmasksoptimizationDRNNs, heymann2016neuralspectralmaskestimation} and ideal affinity matrices \cite{hershey2016deepclustering, Isik2016MultiSpeakerDCL}. Other works have supplemented this by additionally reconstructing the phase of the sources \cite{wang2019deepPhaseReconSpeakerSep, wichern2018phaseReconsLearnedTFrepresentations}. However, the ideal STFT masks impose an upper bound on the separation performance the aforementioned criteria do not necessarily translate to optimal separation. In order to address this, recent works have proposed end-to-end separation schemes where the encoder, decoder and separation modules are jointly optimized using a time-domain loss between the reconstructed sources waveforms and their clean targets \cite{luo2019convTasNet, venkataramani2018end, kavalerov2019universal}. However, a joint time-domain end-to-end training approach might not always yield an optimal decomposition of the input mixtures resulting to worse performance than the fixed STFT bases \cite{kavalerov2019universal}.  
% In doing so, these approaches find a latent space which is fine-tuned towards decomposing the input mixtures.
% these approaches find a latent space for the decomposition of the input mixtures, there is no explicit control .

Some studies have reported significant benefits when performing source-separation in two stages. In \cite{grais2017twostageInterference}, first the sources are separated and in a second stage the interference between the estimated sources is reduced. Similarly, an iterative scheme is proposed in \cite{kavalerov2019universal}, where the separation estimates from the first network are used as input to the final separation network. In \cite{liu2019DeepCASA}, speaker separation is performed by first separating frame-level spectral components of speakers and later sequentially grouping them using a clustering network. Lately, state-of-the-art results in most natural language processing tasks have been achieved by pre-training the encoder transformation network \cite{devlin2019bert}.

In this work, we propose a general two-step approach for performing source separation which can be used in any mask-based separation architecture. First we pre-train an encoder and decoder in order to learn a suitable latent representation. In the second step, we train a separation module using as loss the negative permutation invariant \cite{Yu2017PIT} scale invariant SDR (SI-SDR) \cite{le2019sdr} w.r.t. the learned latent representation. Moreover, we prove that for the case that the decoder is a transpose convolutional layer \cite{luo2019convTasNet, kavalerov2019universal}, SI-SDR on the latent space bounds from below time-domain SI-SDR. Our experiments show that by maximizing SI-SDR on the learned latent targets, a consistent performance improvement is achieved across multiple sound separation tasks compared to the time-domain end-to-end training approach when using the exact same model architecture. The SI-SDR upper bound using the learned latent space is also significantly higher than that of STFT-domain masks. Finally, we also observe that the pre-trained encoder representations are also more sparse and structured compared to the joint training approach. 

\section{Two-step source separation}
\label{sec:method}
Assuming a mixture $\textbf{x} \in \mathbb{R}^T$ that consists of $N$ sources $\textbf{s}_1, \cdots, \textbf{s}_N \in \mathbb{R}^T$ with $T$ samples each in the time-domain, we propose to perform source separation in two independent steps: A) We first obtain a latent representation $\textbf{v}_1, \cdots, \textbf{v}_N \in \mathbb{R}^K$ for the source signals and $\textbf{v}_{\mathbf{x}} \in \mathbb{R}^K$ for the input mixture. B) Then, we train a separation module which operates on the latent representation of the mixture $\textbf{v}_{\mathbf{x}}$ and is trained to estimate the latent representation of the clean sources $\textbf{v}_i$ (or their masks $\textbf{m}_i$ in that space).
\begin{figure}[!htb]
    \centering
  \begin{subfigure}[h]{\linewidth}
      \includegraphics[width=\linewidth]{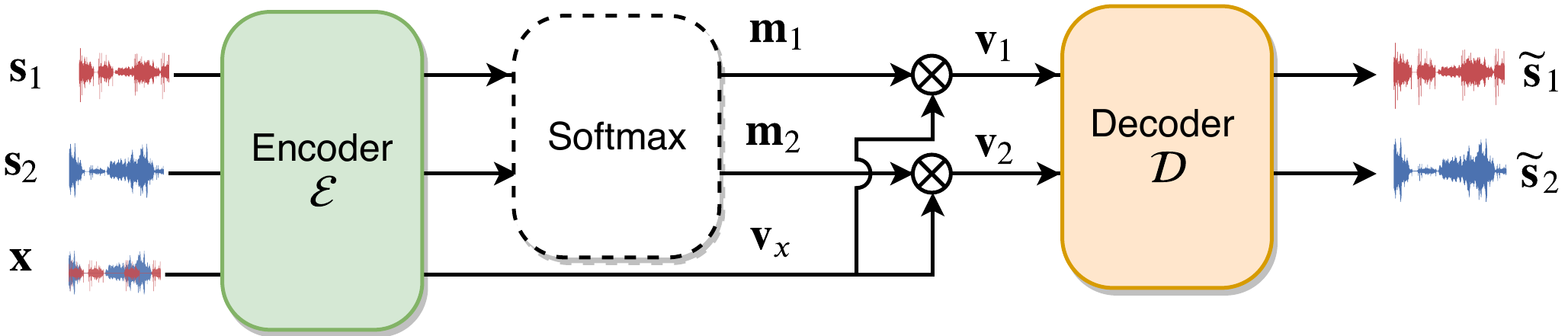}
      \caption{Step 1: Learning the latent targets.}
      \label{fig:step1}
  \end{subfigure} \\
  \begin{subfigure}[h]{\linewidth}
      \includegraphics[width=\linewidth]{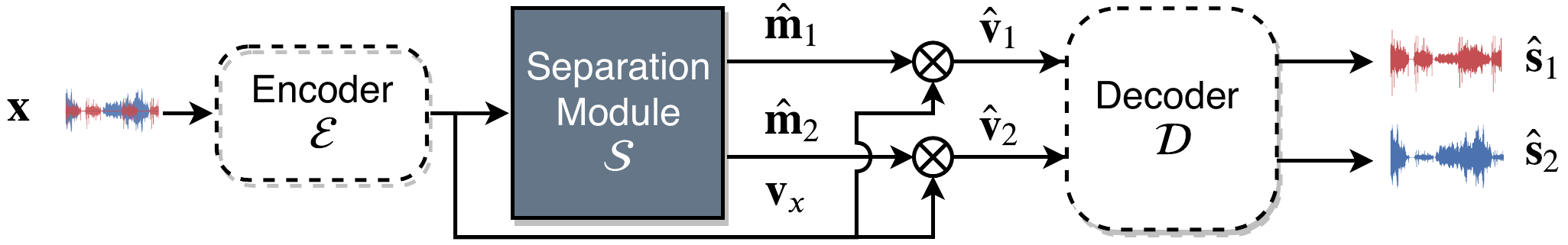}
      \caption{Step 2: Training the separation module only.}
      \label{fig:step2}
  \end{subfigure} 
    \caption{Training a separation network in two independent steps. For each step, the non-trainable parts are represented with a dashed line.}
    \label{fig:twosteps}
    % \vspace{-3pt}
\end{figure} 
\subsection{Step 1: Learning the Latent Targets}
\label{sec:method:step1}
As a first step we train an encoder $\mathcal{E}$ in order to obtain a latent representation for the mixture $\textbf{v}_{\textbf{x}} = \mathcal{E} \left( \textbf{x} \right)$. We also provide the clean sources as inputs to this encoder to obtain $\textbf{v}_1, \cdots, \textbf{v}_N$ and apply a softmax function (across the dimension of the sources) in order to obtain separation masks $\textbf{m}_1, \cdots, \textbf{m}_N$ for each source. An element-wise multiplication of these masks with the latent representation of the mixture  $\textbf{v}_i = \textbf{m}_i \odot \textbf{v}_{\textbf{x}}, \enskip \forall i \in \{1, \cdots, N\}$, can be used as an estimate for each source. The decoder module $\mathcal{D}$ is then trained to transform these latent representations back to time-domain using $\widetilde{\textbf{s}}_i = \mathcal{D} \left(\textbf{v}_{\textbf{i}} \right), \enskip \forall i \in \{1, \cdots, N\}$. In order to train the encoder and the decoder we optimize the permutation invariant \cite{Yu2017PIT} SI-SDR \cite{le2019sdr} between the clean sources $\textbf{s}$ and the estimated sources $\widetilde{\textbf{s}}$:
\begin{equation}
\label{eq:step1SISDR}
    \begin{gathered}
    \mathcal{L}_1 = - \text{SI-SDR}(\textbf{s}^*, \widetilde{\textbf{s}}) = - 10 \log_{10} \left(\| \alpha \textbf{s}^*\|^2 / \| \alpha \textbf{s}^* - \widetilde{\textbf{s}}\|^2 \right)
    \end{gathered}
\end{equation}
where $\textbf{s}^*$ denotes the permutation of the sources that maximizes SI-SDR and the scalar $\alpha =  \widetilde{\textbf{s}}^\top  \textbf{s}^* /\|\textbf{s}\|^2$ ensures that the loss is scale invariant. A schematic representation of the aforementioned step for two sources is depicted in Fig. \ref{fig:step1}. The objective of this step is to find a latent representation transformation, which facilitates source separation through masking.
% \subsubsection{Optimization Using the Clean Sources}
% \subsubsection{Separation Targets on the Latent Space}
\subsection{Step 2: Training the Separation Module}
\label{sec:method:step2}
Once the weights of the encoder and decoder modules are fixed using the training recipe described in Step 1, we can train a separation module $\mathcal{S}$. Given the latent representation of an input mixture $\textbf{v}_{\textbf{x}} = \mathcal{E} \left( \textbf{x} \right)$, $\mathcal{S}$ is trained to produce an estimate of the latent representation $\hat{\textbf{v}}_{i}$ for each clean source $\textbf{v}_{i}$, i.e. $\hat{\textbf{v}} = \mathcal{S} \left( \textbf{v}_{\textbf{x}} \right) $. During inference, we can use the pre-trained decoder to transform the source estimates back into the time-domain $\hat{\textbf{s}} = \mathcal{D} \left( \mathcal{S} \left( \textbf{v}_{\textbf{x}} \right) \right)$. The block diagram describing the training of the separation module with a fixed encoder and decoder is shown in Fig. \ref{fig:step2}.
\subsubsection{Training using SI-SDR on the Latent Separation Targets}
\label{sec:method:step2:SISDRonlatent}
In contrast to recent time-domain source-separation approaches \cite{luo2018tasnet, luo2019convTasNet} which train all modules $\mathcal{E}$, $\mathcal{D}$, and $\mathcal{S}$ using a variant of the loss defined in Eq. \ref{eq:step1SISDR}, we propose to use the permutation invariant SI-SDR directly on the latent representation. For simplicity of notation we assume that each source has a vector latent representation $\textbf{v}_i \in \mathbb{R}^K$ in a high dimensional space. The loss for training the separation module could then be: $\mathcal{L}_2 = - \text{SI-SDR}(\textbf{v}^*, \hat{\textbf{v}}) $. The exact same training procedure could be followed, but now we can use as targets the optimal separation targets on the latent space as opposed to the time domain signals. The premise is that if the separation module is trained on producing latent representations $\hat{\textbf{v}} \approx \textbf{v}$ which are close to the ideal ones (assuming ideal permutation order) then the estimates of the sources after the decoding layer would also approximate the clean sources in time-domain $\hat{\textbf{s}} = \mathcal{D} \left( \mathcal{S} \left( \textbf{v}_{\textbf{x}} \right) \right) \approx  \mathcal{D} \left(\textbf{v} \right) = \widetilde{\textbf{s}} \approx \textbf{s}$. The latter might not hold for any arbitrary embedding process, but in the next section we prove that SI-SDR in the latent representations lower-bounds the SI-SDR in the time-domain. 
\subsubsection{Relation to maximization of SI-SDR on Time-Domain}
\label{sec:method:step2:proof}
We restrict ourselves to a decoder that consists of a 1-D transposed convolutional layer which is the same as the decoder selection in most of the current end-to-end source separation approaches \cite{luo2018tasnet, luo2019convTasNet, kavalerov2019universal, venkataramani2018end}. For this part we focus on the $i$th target latent representation $\textbf{v}_i \in \mathbb{R}^K$ that corresponds to a source time-domain signal $\textbf{s}_i \in \mathbb{R}^T$. Because the encoder-decoder modules are trained as described in Section \ref{sec:method:step1}, the separation target produced by the auto-encoder $\widetilde{\textbf{s}}_i$ would be close to the clean source $\textbf{s}_i$, namely:
\begin{equation}
\label{eq:clean_source_decoder}
    \mathcal{D}  \left( \textbf{v}_{\textbf{i}} \right) = \widetilde{\textbf{s}}_i  \approx \textbf{s}_i
\end{equation}
The separation network produces an estimated latent vector $\hat{\textbf{v}}_i$ that corresponds to an estimated time-domain signal $\hat{\textbf{s}}_i = \mathcal{D} \left( \hat{\textbf{v}}_i\right)$. Because the decoder is just a convolutional layer we can express it as a linear projection $\mathcal{D}: \mathbb{R}^K \rightarrow \mathbb{R}^T$ using the matrix $\textbf{P} \in \mathbb{R}^{T \times K}$:
\begin{equation}
\label{eq:decoder_equalities}
    \hat{\textbf{s}}_i = \textbf{P} \hat{\textbf{v}}_i, \enskip \widetilde{\textbf{s}}_i = \textbf{P} \textbf{v}_i, \enskip \forall i \in \{1, \cdots, N\}
\end{equation}
Assuming the Moore-Penrose pseudo-inverse of $\textbf{P}$ is well defined, we express the inverse mapping from time to the latent-space as:
\begin{equation}
\label{eq:moore_penrose}
    \hat{\textbf{v}}_i = \textbf{P}^\dagger  \hat{\textbf{s}}_i, \enskip   \textbf{v}_i = \textbf{P}^\dagger  \widetilde{\textbf{s}}_i, \enskip \forall i \in \{1, \cdots, N\} 
\end{equation}
\begin{prop}
\label{lem:innerprod}
Let $\textbf{y}, \hat{\textbf{y}} \in \mathbb{R}^d$ and their corresponding projections through $\textbf{A} \in \mathbb{R}^{n \times d}$ to $\mathbb{R}^n$ defined as  $\textbf{A} \textbf{y}$ and $\textbf{A}  \hat{\textbf{y}}$, respectively. If $\| \textbf{y}\| = \| \hat{\textbf{y}} \| = 1$ then the absolute value of their inner product on the projection space $\mathbb{R}^n$ is bounded above from the absolute value of their inner product in $\mathbb{R}^d$, namely: $\left( \hat{\textbf{y}}^\top \textbf{A}^\top \textbf{A} \textbf{y} \right)^2 \leq g\left ( \textbf{A} \right) + \left( \hat{\textbf{y}}^\top \textbf{y} \right)^2$, where $g\left ( \textbf{A} \right) \geq 0$ and depends only on the values of $\textbf{A}$. 
% \vspace{-1pt}
\begin{proof}
The inner product in the projection space can be rewritten as:
\begin{equation}
\label{eq:squareexpand}
\begin{gathered}
\left( \hat{\textbf{y}}^\top \textbf{A}^\top \textbf{A} \textbf{y} \right)^2  = \left[ \hat{\textbf{y}}^\top \left( \textbf{A}^\top \textbf{A} - \textbf{I}   \right)\textbf{y} + \hat{\textbf{y}}^\top\textbf{y} \right]^2 = \\ 
\left[ \hat{\textbf{y}}^\top \left( \textbf{A}^\top \textbf{A} - \textbf{I}   \right)\textbf{y} \right]^2 +
2 \left[ \hat{\textbf{y}}^\top \left( \textbf{A}^\top \textbf{A} - \textbf{I}   \right)\textbf{y}  \hat{\textbf{y}}^\top\textbf{y} \right] + \left(\hat{\textbf{y}}^\top\textbf{y} \right)^2 
\end{gathered}
\end{equation}
Moreover, we can bound the first term of Eq. \ref{eq:squareexpand} by applying Cauchy-Schwarz inequality to the inner products and using the fact that $\| \textbf{y}\| = \| \hat{\textbf{y}} \| = 1$ as shown next:
\begin{equation}
\label{eq:first_part_bound}
\begin{gathered}
\left[ \hat{\textbf{y}}^\top \left( \textbf{A}^\top \textbf{A} - \textbf{I}   \right)\textbf{y} \right] \leq \| \hat{\textbf{y}} \| \cdot  \| \textbf{A}^\top \textbf{A} - \textbf{I} \| \cdot \| \textbf{y} \| = \| \textbf{A}^\top \textbf{A} - \textbf{I} \|
\end{gathered}
\end{equation}
Similarly, we use Cauchy-Schwarz inequality and inequality \ref{eq:first_part_bound} in order to bound the second term of Eq. \ref{eq:squareexpand} as well:
\begin{equation}
\label{eq:second_part_bound}
\begin{gathered}
\left[ \hat{\textbf{y}}^\top \left( \textbf{A}^\top \textbf{A} - \textbf{I}   \right)\textbf{y}  \hat{\textbf{y}}^\top\textbf{y} \right] \leq  \| \textbf{A}^\top \textbf{A} - \textbf{I} \|
\end{gathered}
\end{equation}
Then by applying inequalities \ref{eq:first_part_bound} and \ref{eq:second_part_bound} to Eq. \ref{eq:squareexpand} we get:
\begin{equation}
\label{eq:final_concl_inequality}
\begin{split}
\left( \hat{\textbf{y}}^\top \textbf{A}^\top \textbf{A} \textbf{y} \right)^2 & \leq \| \textbf{A}^\top \textbf{A} - \textbf{I} \|^2 + 2\cdot \| \textbf{A}^\top \textbf{A} - \textbf{I} \| + \left( \hat{\textbf{y}}^\top \textbf{y} \right)^2 
\end{split}
\end{equation}
where always $g\left ( \textbf{A} \right) = \| \textbf{A}^\top \textbf{A} - \textbf{I} \|^2 + 2\cdot \| \textbf{A}^\top \textbf{A} - \textbf{I} \| \geq 0$. Finally, we conclude that $\left( \hat{\textbf{y}}^\top \textbf{A}^\top \textbf{A} \textbf{y} \right)^2 \leq g\left ( \textbf{A} \right) + \left( \hat{\textbf{y}}^\top \textbf{y} \right)^2$. 
\end{proof}
\end{prop} 
\begin{prop}
\label{lem:sisdr}
Let $\textbf{y}, \hat{\textbf{y}} \in \mathbb{R}^d$, with unit norms, then maximizing $\text{SI-SDR}(\textbf{y}, \hat{\textbf{y}})$ w.r.t. $\hat{\textbf{y}}$ is equivalent to maximizing $  \left( \hat{\textbf{y}}^\top \textbf{y} \right)^2$  w.r.t. $\hat{\textbf{y}}$.
\begin{proof}
By assuming that there is an optimal solution $\hat{\textbf{y}}^\star$:
\begin{equation}
\begin{split}
\hat{\textbf{y}}^\star  & =
 \underset{\hat{\textbf{y}}}{\operatorname{\arg \max}} \enskip  \text{SI-SDR}(\textbf{y}, \hat{\textbf{y}}) \xlongequal{ Eq. \ref{eq:step1SISDR}} \underset{\hat{\textbf{y}}}{\operatorname{\arg \max}} \enskip  \frac{\| \alpha \textbf{y}\|^2}{\| \alpha \textbf{y} - \hat{\textbf{y}}\|^2}  = \\ & =
\underset{\hat{\textbf{y}}}{\operatorname{\arg \max}} \enskip \frac{\| \alpha \textbf{y}\|^2}{
\| \alpha  \textbf{y}\|^2 + \|  \hat{\textbf{y}}\|^2 - 2 \alpha \hat{\textbf{y}}^\top \textbf{y}} =
\\ & = \underset{\hat{\textbf{y}}}{\operatorname{\arg \max}} 
\left[ 1 + \frac{\|  \hat{\textbf{y}}\|^2}{\|\textbf{y}\|^2 \alpha^2 } - 2 \frac{  \hat{\textbf{y}}^\top \textbf{y}}{\alpha \|\textbf{y} \|^2} \right]^{-1} \xlongequal{\alpha = \nicefrac{\hat{\textbf{y}}^\top \textbf{y}}{\|\textbf{y} \|^2}} \\
& = \underset{\hat{\textbf{y}}}{\operatorname{\arg \max}} 
\frac{ \left( \hat{\textbf{y}}^\top \textbf{y} \right)^ 2}{ \|\textbf{y} \|^2 \|\hat{\textbf{y}} \|^2 }   \xlongequal{\|\textbf{y} \|=\|\hat{\textbf{y}} \|=1} \underset{\hat{\textbf{y}}}{\operatorname{\arg \max}} \left( \hat{\textbf{y}}^\top \textbf{y} \right)^2 
\end{split} \end{equation}Which means that the two optimization goals are equivalent. \end{proof}  \end{prop} 
\noindent Now we focus on the relationship of the maximization of SI-SDR for the $i$th source when it is performed directly on the latent space $\text{SI-SDR}(\textbf{v}_i, \hat{\textbf{v}}_i)$ and when it is performed on the time-domain using the clean source as a target $\text{SI-SDR}(\textbf{s}_i, \hat{\textbf{s}}_i) \approx \text{SI-SDR}(\widetilde{\textbf{s}}_i, \hat{\textbf{s}}_i)$. Again because all the SI-SDR measures are scale-invariant, we can assume that the separation targets and the estimates vectors have unit norms on both the time-domain and the latent space, namely $\|\hat{\textbf{v}}_i \| = \|\textbf{v}_i \|= \| \hat{\textbf{s}}_i\| = \|\widetilde{\textbf{s}}_i \| = 1$. By using Proposition \ref{lem:innerprod} we get:
\begin{equation}
    \left(\hat{\textbf{v}}_i^\top \textbf{v}_i\right) ^2 = \left[\hat{\textbf{s}}_i^\top \left( \textbf{P}^\dagger \right)^\top \textbf{P}^\dagger \widetilde{\textbf{s}}_i\right] ^2  \leq g\left ( \textbf{P}^\dagger \right) + \left(\hat{\textbf{s}}_i^\top  \widetilde{\textbf{s}}_i\right) ^2 
\end{equation}
Thus, by using the auto-encoder property (Eq. \ref{eq:clean_source_decoder}) and Proposition \ref{lem:sisdr} we conclude that $\text{SI-SDR}(\textbf{v}_i, \hat{\textbf{v}}_i)$ on the latent space lower bounds the corresponding value $\text{SI-SDR}(\textbf{s}_i, \hat{\textbf{s}}_i)$ on the time-domain. The same proof holds for any encoder $\mathcal{E}$ and for other targets on the latent space such as the masks $\textbf{m}_i \in [0,1]^K$. Empirically, we indeed notice that the maximization of  $\text{SI-SDR}(\textbf{v}_i, \hat{\textbf{v}}_i)$ on the latent space leads to the maximization of $\text{SI-SDR}(\textbf{s}_i, \hat{\textbf{s}}_i)$ on the time-domain. 

\section{Experimental Framework} \label{sec:Framework} 
To experimentally verify our approach we perform a set of source 
separation experiments as described in the following sections. 
\subsection{Audio Data}
We use two audio data collections. For speech sources we use $14,823$ speech utterances from Wall street journal (WSJ0) corpus \cite{WSJ0}. Training, validation and test speaker mixtures are generated by randomly selecting various speakers from the sets {\fontfamily{qcr}\selectfont si\_tr\_s}, {\fontfamily{qcr}\selectfont si\_dt\_05} and {\fontfamily{qcr}\selectfont si\_et\_05}, respectively. % We use audio clips which are at least $4$secs long.% \vspace*{1mm}

For non-speech sounds we use the $2,000$ $5$secs audio clips which are equally balanced between $50$ classes from the environmental sound classification (ESC50) data collection \cite{esc50}. ESC50 spans various sound categories such as: \textit{non-speech human sounds}, \textit{animal sounds}, \textit{natural soundscapes}, \textit{interior sounds} and \textit{urban noises}. We split the data to train, validation and test sets with a ratio of $8:1:1$, respectively. For each set, the same prior is used across classes (e.g., each class has the same number of clips). Also, the sets do not share clips which originate from the same initial source file.  
\subsection{Sound Source Separation Tasks}
\label{sec:Framework:SeparationTasks}
In order to develop a system capable of performing universal sound source separation \cite{kavalerov2019universal}, we evaluate our two-step approach under three distinct sound separation tasks. For all separation tasks, each input mixture consists of two sources which are always mixed using $4$secs of their total duration. All audio clips are downsampled to $8$kHz for efficient processing. We discuss the audio collection(s) that we utilize and the mixture generation process in the sections below.
\subsubsection{Speech Separation} \label{sec:Framework:SeparationTasks:WSJ}
We only use audio clips containing human speech from WSJ0. In accordance to other studies performing experiments on single-channel speech source separation \cite{luo2019convTasNet, shi2019furcax, le2019phasebook, wichern2018phaseReconsLearnedTFrepresentations, wang2019deepPhaseReconSpeakerSep}, we use the publicly available WSJ0-2mix dataset \cite{hershey2016deepclustering}. In total there are $20,000$, $5,000$ and $3,000$ mixtures for training, validation and testing, correspondingly.
\subsubsection{Non-Speech Separation} \label{sec:Framework:SeparationTasks:ESC50}
We use audio clips only from ESC50. In this case, the total number of the available clean sources sounds is small, and thus, we propose an augmented mixture generation process which enables the generation of much more diverse mixtures. In order to generate each mixture, we randomly select a $4$sec segment from two audio files from two distinct audio classes. We mix these two segments with a random signal to noise ratio (SNR)s between $-2.5$ and $2.5$dB. For each epoch, $20,000$ training mixtures are generated which generally are not the same with the ones generated for other epochs. For validation and test sets we fix their random seeds in order to always evaluate on the same $5,000$ and $3,000$ generated mixtures, respectively.
\subsubsection{Mixed Separation} \label{sec:Framework:SeparationTasks:WSJESC50}
All four possible mixture combinations between speech and non-speech audio are considered by using both WSJ0 and ESC50 sources. Building upon the data augmentation training idea, we also add a random variable which controls the data collection (ESC50 or WSJ0) from which a source waveform is going to be chosen. Specifically, we set an equal probability of choosing a source file from the two collections (ESC50 and WSJ0). For WSJ0 each speaker is considered a distinct sound class, thus, no mixture consists utterances from the same speaker. After the two source waveforms are chosen, we follow the mixture generation process described in Section \ref{sec:Framework:SeparationTasks:ESC50}.  
\subsection{Selected Network Architectures}
Based on recent state-of-the-art approaches on both speech and universal sound source separation with learnable encoder and decoder modules, we consider configurations for the encoder-decoder parts as well as the separation module which are based on a similar time-dilated convolutional network (TDCN) architecture. In particular, we consider our implementations of ConvTasNet \cite{luo2019convTasNet} that we refer simply as TDCN and its improved version proposed in \cite{kavalerov2019universal} that we refer as residual-TDCN (RTDCN). 
\subsubsection{Encoder-Decoder Architecture}
The encoder $\mathcal{E}$ consists of one $1$D convolutional layer and a ReLU activation on top in order to ensure a non-negative latent representation of each audio input. Following the assumptions stated in Section \ref{sec:method:step2:proof}, we use a $1$D transposed convolutional layer for the decoder $\mathcal{D}$. Both encoder and decoder have the same number of channels (or number of bases) and their $1$D kernels have a length corresponding to $2.625$ms ($21$ samples) and a hop-size equivalent to $1.25$ms ($10$ samples). For each task we select a different number of channels for the encoder and the decoder modules ($32$, $128$ and $256$ for speech only, mixed and non-speech only separation tasks, respectively). 
\subsubsection{Separation Modules Architectures}
Our implementation of TDCN consists of the same architecture and parameter configuration for the separation module as described in \cite{luo2019convTasNet} with an additional batch normalization layer before the final mask estimation which improved its performance over the original version on all separation tasks. Inspired by the original RTDCN separation module \cite{kavalerov2019universal}, we keep the same parameter configuration as TDCN and we additionally use a feature-wise normalization between layers instead of global normalization. We also add long-term residual connections from previous layers. Moreover, before summing the residual connections, we concatenate them, normalize them and feed them through a dense layer as the latter yields some further improvement in separation performance. (Code is available online\footnote{Source code: \href{https://github.com/etzinis/two_step_mask_learning}{github.com/etzinis/two\_step\_mask\_learning}}.)

\subsection{Training and Evaluation Details}
In order to show the effectiveness of our proposed two-step approach, we use the same network architecture when we perform end-to-end time-domain source separation and use as a loss the negative SI-SDR between the estimated signals on the time-domain and the clean waveforms $-\text{SI-SDR}(\hat{\textbf{s}}, \textbf{s}^*)$. Instead in our two-step approach, we train the encoder-decoder parts separately as described in Section \ref{sec:method:step1}. In the second step, we use the pre-trained encoders for each task and train the separation module using as loss the negative SI-SDR on the latent space targets $-\text{SI-SDR}(\textbf{v}^*, \hat{\textbf{v}})$ or their corresponding masks $-\text{SI-SDR}(\textbf{m}^*, \hat{\textbf{m}})$ (see Section \ref{sec:method:step2}). We train all models using the Adam optimizer \cite{adam}, the batch size is equal to $4$, the initial learning rate is set to $0.001$ and we decrease it by a factor of $10$ at the $100$th epoch. We train TDCN and RTDCN separation networks for $100$ epochs and $120$ epochs, respectively. The encoder-decoder parts for each task are trained independently for $200$ epochs ($100$ times faster than training the separation network). We evaluate the separation performance for all models using SI-SDR improvement (SI-SDRi) on time domain which is the difference of SI-SDR of the estimated signal and the input mixture signal \cite{luo2019convTasNet, kavalerov2019universal}. As the STFT oracle mask we choose the ideal ratio mask (IRM) using a Hanning window with $64$ms length and $16$ms hop-size \cite{luo2019convTasNet}.  
\section{Results \& Discussion}
\subsection{Comparison with Time-Domain Separation}
In Table \ref{tab:best_models}, the mean separation performance of best models is reported for each task. We notice that the proposed two-step approach and training on the latent space leads to a consistent improvement over the end-to-end approach where we train the same architecture using the time-domain SI-SDR loss. This observation holds when different separation modules are used and when we test them under different separation tasks. The non-speech separation task seems the hardest one since the models have access to only a limited number of training mixtures which further underlines the importance of our proposed data-augmentation technique as described in Section \ref{sec:Framework:SeparationTasks:ESC50}. Our two-step approach yields an absolute SI-SDR improvement over the end-to-end baseline of up to $0.7$dB, $0.5$dB and $0.7$dB for speech, non-speech and mixed separation tasks, respectively. Notably, this performance improvement is achieved using the exact same architecture but instead of training it end-to-end using a time-domain loss, we pre-train the auto-encoder part and use a loss on the latent representations of the sources. 
\begin{table}[!htb]
    \centering
    \begin{tabular}{c|c|c|c|c}
    \hlinewd{1pt}
    \toprule
    Separation  & Target  & \multicolumn{3}{c}{Sound Separation Task} \\ \cline{3-5}
    Module & Domain & Speech & Non-speech & Mixed \\
    \hlinewd{1pt}
    \multirow{2}{*}{TDCN} & Time & 15.4 & 7.7 & 11.7 \\ \cline{2-5}
     & Latent & 16.1 & 8.2 & 12.4 \\ \hlinewd{1pt}
     \multirow{2}{*}{RTDCN} & Time & 15.6 & 8.3 & 12.0 \\ \cline{2-5}
     & Latent & 16.2 & 8.4 & 12.6 \\ \hlinewd{1pt}
     Oracle & STFT & 13.0 & 14.8 & 14.5 \\ \cline{2-5}
    Masks & Latent & 34.1 & 39.2 & 39.5 \\
    \bottomrule
    \end{tabular}
    \caption{Mean SI-SDRi (dB) of best performing models.}
    \label{tab:best_models}
    % \vspace{-5pt}
\end{table}
\begin{figure}[!htb]
    \centering
  \begin{subfigure}[h]{\linewidth}
      \includegraphics[width=\linewidth]{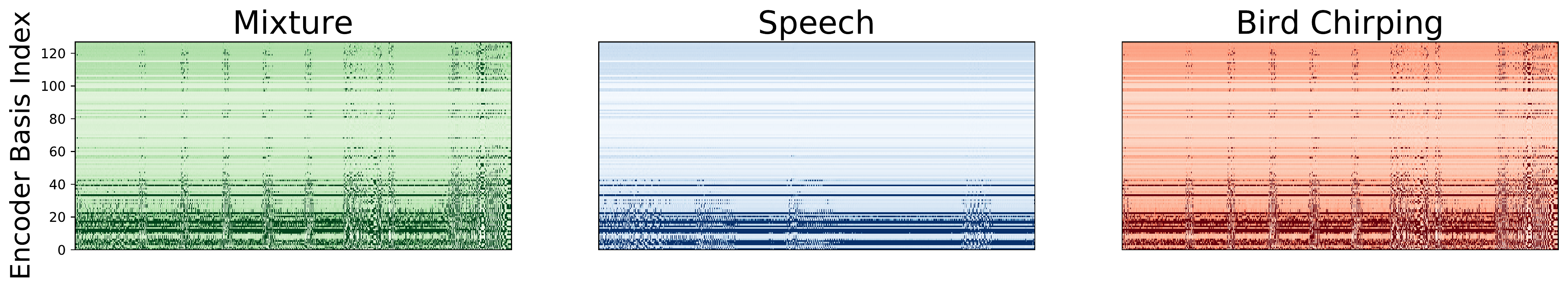}
      \caption{Two-step training approach.}
      \label{fig:our_latents}
  \end{subfigure} \\
  \begin{subfigure}[h]{\linewidth}
      \includegraphics[width=\linewidth]{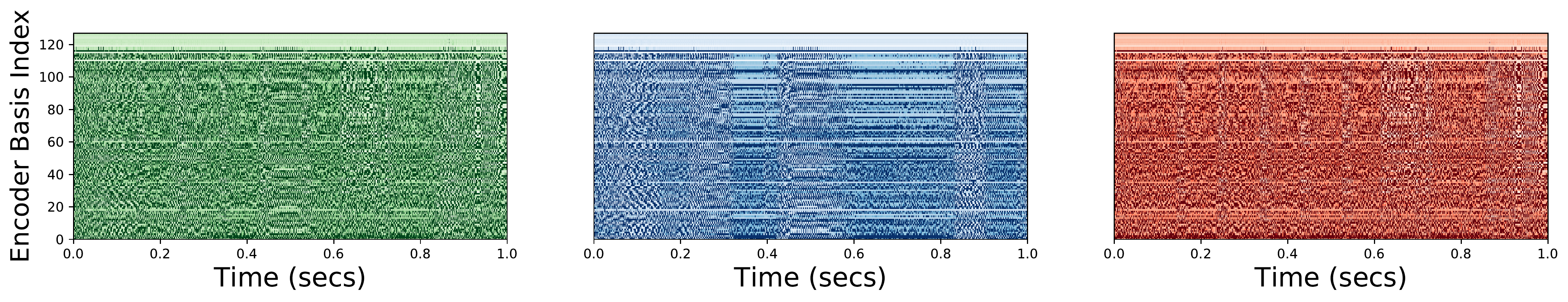}
      \caption{Joint end-to-end training using time-domain SI-SDR loss.}
      \label{fig:tasnet_latents}
  \end{subfigure} 
    \caption{Latent representations of a $1$sec mixture and its constituent sources when training the same encoder architecture: a) individually using the proposed two-step approach (top) b) jointly with the TDCN separation module using SI-SDR loss on time-domain (bottom). We sort the basis indexes w.r.t. their energy and we raise the value of each cell to $0.1$ for better visualization.}
    \label{fig:latentrepresentations}
    % \vspace{-8.8pt}
\end{figure} 
\subsection{Separation Targets in the Latent Space}
In Table \ref{tab:best_models}, we see that the oracle mask obtained from the two-step approach gives a much higher upper bound of separation performance, for all tasks, compared to ideal masks on the STFT domain. This is in line with the prior work that proposed to decompose signals using learned transforms \cite{luo2019convTasNet, venkataramani2018end}. In Fig. \ref{fig:latentrepresentations} we can qualitatively compare the latent representations obtained from the same encoder when trained with our proposed two-step approach and with the baseline joint training of all modules. When the encoder and decoder are trained individually, a fewer number of bases are used to encode the input which leads to a sparser representation ($\ell_1$ norm is roughly $10\times$ smaller compared to the joint training approach). Finally, the latent representations obtained from our proposed approach exhibit a spectrogram-like structure in a way that \textit{Speech} is encoded using less bases than high frequency sounds like \textit{Bird Chirping}. 
\section{Conclusion}
We show how by pre-learning an optimal latent space can result in better source separation performance compared to a time-domain end-to-end training approach. Our experiments show that the proposed two-step approach yields a consistent performance improvement under multiple sound separation tasks. Additionally, the obtained sound latent representations remain sparse and structured while they also enjoy a much higher upper bound of separation performance compared to STFT-domain masks. Although this approach was demonstrated on TDCN architectures, it can be easily adapted for use with any other mask-based system. %A promising future direction is to combine the proposed latent-space SI-SDR loss with a wider family of architectures that use custom latent spaces. 
% The use of the SI-SDR in the latent space is also applicable to a wider family of architectures that use custom latent spaces.
\vfill\pagebreak \balance
\bibliographystyle{IEEEbib}
\bibliography{refs}
\end{document}